\pdfoutput=1

\documentclass[11pt]{article}

\usepackage[final]{acl}

\usepackage{times}
\usepackage{latexsym}
\usepackage{tikz}
\usepackage{makecell}
\usepackage{amssymb}
\usepackage{pifont}
\usepackage{subcaption}
\usepackage{amsmath}
\usepackage{multirow}
\usepackage{booktabs}
\usepackage{bigstrut}

\usepackage[T1]{fontenc}

\usepackage[utf8]{inputenc}

\usepackage{microtype}

\usepackage{inconsolata}

\usepackage{graphicx}
\usepackage{titlesec}

\newcommand{\zeyu}[1]{\textcolor{black}{#1}}
\newcommand{\revise}[1]{\textcolor{black}{#1}}

\title{MemBench: Towards More Comprehensive Evaluation on the Memory of LLM-based Agents}

\author{
 \textbf{Haoran Tan\textsuperscript{1}\footnotemark[1]\footnotemark[3]\footnotemark[4]},
 \textbf{Zeyu Zhang\textsuperscript{1}\footnotemark[1]\footnotemark[3]\footnotemark[4]},
 \textbf{Chen Ma\textsuperscript{1}\footnotemark[3]\footnotemark[4]},
 \textbf{Xu Chen\textsuperscript{1}\footnotemark[2]\footnotemark[3]\footnotemark[4]},
\\
 \textbf{Quanyu Dai\textsuperscript{2}\footnotemark[2]},
 \textbf{Zhenhua Dong\textsuperscript{2}}
\\
\\
 \textsuperscript{1}Gaoling School of Artificial, Renmin University of China, Beijing, China,\\
 \textsuperscript{2}Huawei Noah's Ark Lab
\\
\texttt{\{tanhaoran1321,zeyuzhang,xu.chen\}@ruc.edu.cn,}
\texttt{daiquanyu@huawei.com}
}

\begin{document}
\maketitle
\footnotetext[1]{Co-first authors.}
\footnotetext[2]{Corresponding authors.}
\footnotetext[3]{Beijing Key Laboratory of Research on Large Models and Intelligent Governance}
\footnotetext[4]{Engineering Research Center of Next-Generation Intelligent Search and Recommendation, MOE}

\begin{abstract}

\zeyu{Recent works have highlighted the significance of memory mechanisms in LLM-based agents, which enable them to store observed information and adapt to dynamic environments.}
\zeyu{However, evaluating their memory capabilities still remains challenges.}
\zeyu{Previous evaluations are commonly limited by the diversity of memory levels and interactive scenarios. They also lack comprehensive metrics to reflect the memory capabilities from multiple aspects.}
\zeyu{To address these problems, in this paper, we construct a more comprehensive dataset and benchmark to evaluate the memory capability of LLM-based agents.}
\zeyu{Our dataset incorporates factual memory and reflective memory as different levels, and proposes participation and observation as various interactive scenarios.}
\zeyu{Based on our dataset, we present a benchmark, named MemBench, to evaluate the memory capability of LLM-based agents from multiple aspects, including their effectiveness, efficiency, and capacity. To benefit the research community, we release our dataset and project at  \url{https://github.com/import-myself/Membench}.}

\end{abstract}

\section{Introduction}

In recent years, large language models (LLMs) have demonstrated remarkable capabilities in processing natural languages and performing complex tasks across various domains~\cite{zhao2023survey, wu2024survey}. However, vanilla LLMs typically operate in static scenarios, without interacting with external environments, thereby limiting their potential advancement toward artificial general intelligence (AGI).
To address this limitation, many recent works propose LLM-based agents with extra modules besides the foundation models, enabling them to interact with environments with autonomous learning and dynamic adaptation~\cite{wang2024survey, xi2025rise}.
Among them, the memory module serves as an essential foundation for saving critical information and accumulating experiences. It empowers LLM-based agents to better meet the demands of dynamic tasks, as well as evolve within their environments continuously~\cite{zhang2024survey}.

Some previous studies evaluate the memory capability of LLM-based agents in a subjective way, which adopts human evaluators or LLMs to score the memory process~\cite{zhong2024memorybank}. Other studies focus on the evaluation in an indirect way~\cite{packer2023memgpt}. They measure the task performances of agents conditional on different memory mechanisms, where a better memory mechanism generally leads to better performances.
Recently, some studies introduce long-term dialogue datasets, which can be used to evaluate the long-term memory capabilities of LLM-based agents objectively~\cite{wu2024longmemeval}.

However, previous works have some limitations on evaluating the memory capability of LLM-based agents.
First of all, most of them provide insufficient evaluation of the different levels of memory capabilities, which primarily focus on factual memory while neglecting reflective memory.
Here, we define the factual memory as a low-level type of memory that involves information that is explicitly provided. In contrast, the reflective memory stands a higher level, which is not explicitly stated but can be implicitly reflected. For example, a user's taste preferences represent reflective memory, while their preference for specific dishes is factual memory.
Second, most of them are limited to participation scenarios, where the agent interacts with the user from a first-person perspective. However, in the agent's daily usage, there are also observation scenarios, where the agent observes and records the user's messages from a third-person perspective.
Moreover, most of them are just focusing on the effectiveness of memory mechanisms without considering their efficiency and capacity, which is also significant in real-world applications.

To address these limitations, we propose a more comprehensive dataset and benchmark to evaluate the memory capability of LLM-based agents.
The major features of our dataset and benchmark are presented as follows:

\textbf{Multi-scenario Dataset.} To evaluate the agent's memory capabilities across different scenarios, our dataset includes data from two common usage scenarios. The first is the participation scenario, where the agent interacts with the user. The second is the observation scenario, where the agent is assumed as the role of an observer and required to record the information provided by the user.

\textbf{Multi-level Memory Content.} Our dataset focuses on both factual memory and reflective memory, enabling a comprehensive evaluation of the memory capability of LLM-based agents. It allows for the evaluation of memory capabilities in tasks including information extraction, cross-session reasoning, knowledge updating, temporal reasoning, as well as reflective summarization.

\textbf{Multi-metric Evaluation.} Based on our dataset, we introduce a multi-metric benchmark to evaluate the memory capabilities of LLM-based agents. To provide a comprehensive and obvious assessment of the various aspects of the agent's memory performance, we offer four evaluation metrics, including accuracy, recall, capacity, and temporal efficiency. 

In summary, we introduce a dataset featuring multi-scenario and multi-level content, which is distinctively different from previous datasets. Additionally, we introduce a more comprehensive benchmark with mutli-metric evaluations.
To benefit the research community, we have released our dataset and project at Github repository\footnote{\url{https://github.com/import-myself/Membench}}.
In the following parts, we provide the related works in Section~\ref{sec:related_works}. We illustrate the process of data construction in Section~\ref{sec:dataset_cons}, and present the benchmark with analyses in Section~\ref{sec:benchmark}. Finally, we draw conclusions of our paper in Section~\ref{sec:conclusion}.

\section{Related Works}
\label{sec:related_works}

In recent years, LLM-based agents have been widely applied in many fields, such as recommendation~\cite{wu2024survey}, finance~\cite{ding2024large} and personal assistants~\cite{li2024personal}, because of their great capabilities in solving complex tasks and interactive scenarios~\cite{wang2024survey}. Among the various abilities of agents to solve problems, memory is one of the most important, which is responsible to store observed information and recall relevant experiences, in order to support LLM inference~\cite{zhang2024survey}. The evaluation on the memory capability of LLM-based agents is a critical problem for developing advanced memory.

Previous datasets used for memory evaluation mainly come from dialogue datasets designed to evaluate chat assistants, focusing on assessing the factual memory capabilities of the assistant.
LoCoMo~\citep{maharana2024evaluating} constructs long-term conversations with LLM-expanded personalized descriptions and temporal event graphs, creating conversations for multiple evaluation tasks.
LongMemEval~\citep{wu2024longmemeval} builds a user-assistant interaction dataset with attribute ontology and timestamped history.
\citet{xu2022long} adapts PersonaChat~\citep{zhang2019neural} through translation and role-playing, with annotated personalization usage and partial information visibility. 
\citet{longbench} propose a bilingual benchmark in English and Chinese, covering comprehensive tasks like Q\&A, summarization, and code completion.
\citet{l_eval} provide a dataset designed to evaluate long-context language models across diverse domains and input lengths.
PerLTQA~\cite{du2024perltqa} generates character profiles and events with ChatGPT and Wikipedia, creating profiles, events, and QA pairs after manual validation.

\begin{table}[t]
  \centering
  \caption{The comparision among different datasets. PS indicates Participation Scenario. OS indicates Observation Scenario. FM indicates Factual Memory. RM indicates Reflective Memory.}
  \vspace{-0.1cm}
  \resizebox{\linewidth}{!}{
    \begin{tabular}{>{\centering\arraybackslash}p{2.0cm}>{\centering\arraybackslash}p{1.6cm}>{\centering\arraybackslash}p{2.0cm}>{\centering\arraybackslash}p{2.0cm}}
\hline
\hline
Datasets & Profiles & Scenarios & Levels \bigstrut\\
\hline
PerLTQA & $\checkmark$ & PS    & FM \bigstrut[t]\\
LoCoMo & $\times$ & PS    & FM \\
LongMemEval & $\times$ & PS    & FM \\
MemBench & $\checkmark$ & PS \& OS & FM \& RM \bigstrut[b]\\
\hline
\hline
    \end{tabular}%
    }
  \label{tab:comp}
  \vspace{-0.5cm}
\end{table}%

As we have shown in Table~\ref{tab:comp}, most of these datasets lack diverse scenarios, focusing only on participation scenarios (PS) and overlooking the agent's observation scenarios (OS). Additionally, they only focus on factual memory (FM), neglecting reflective memory (RM). Some works do not include the profiles of users. Previous studies typically use these datasets in long-context evaluation methods that do not align with the agent's memory process. Moreover, the evaluation metrics used with these datasets are not comprehensive.

Compared with previous works, our work is the first study that emphasizes reflective memory, puts forward observation scenarios, adopts evaluation methods that are better suited to the agent's memory process, with more comprehensive metrics.

\begin{figure*}[t]
  \includegraphics[width=\textwidth]{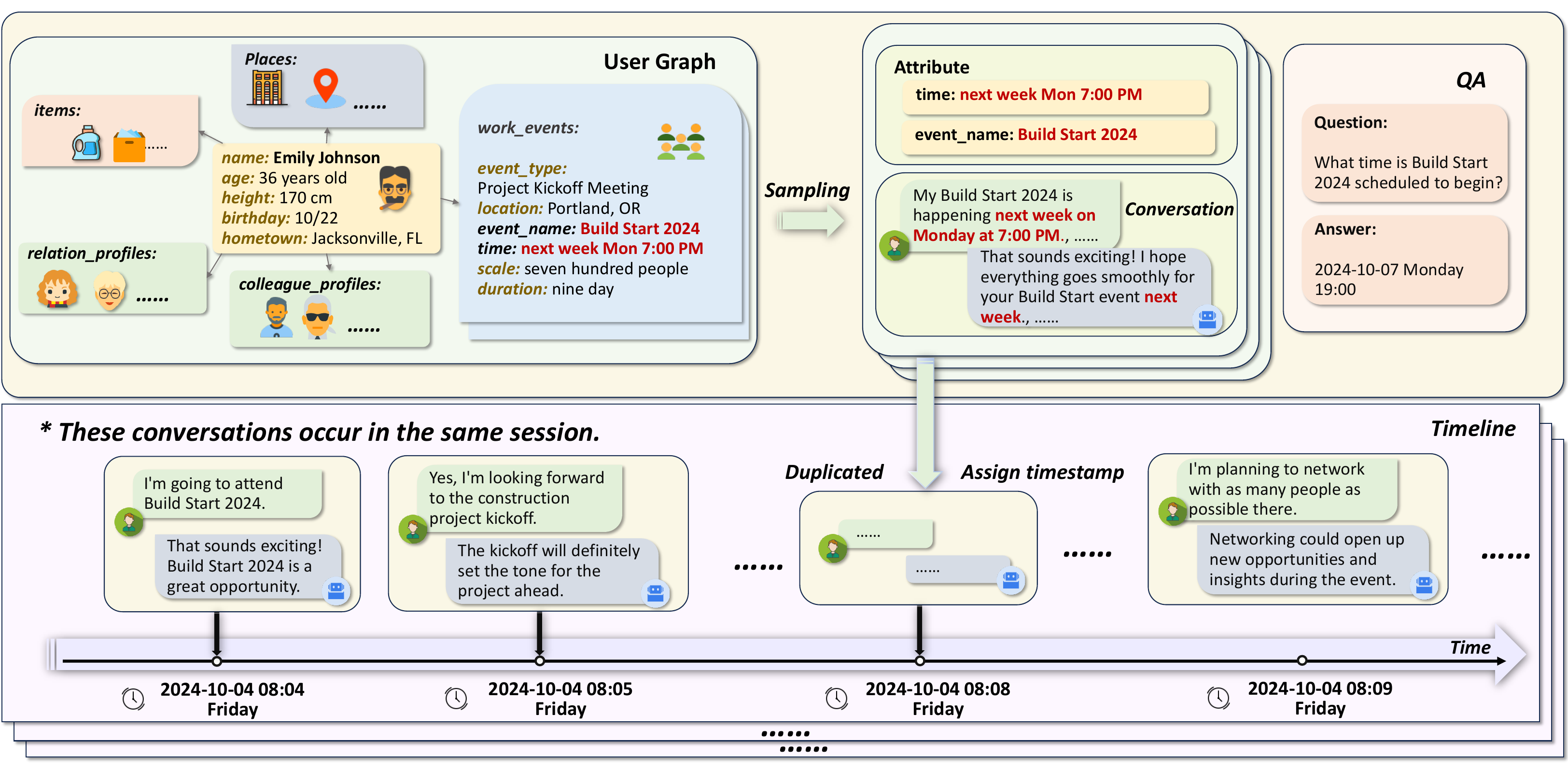}
  \vspace{-0.6cm}
  \caption{An example of generating dialogue data. First, the event "Build Start 2024" is extracted with the time "next week Mon 7:00 PM," which is then used to generate evidence dialogues and questions. It's merged with dialogues generated from other attributes to form a complete dialogue, and an answer is generated based on the provided time label "2024-10-07 Monday 19:00".}
  \vspace{-0.4cm}
  \label{fig:generate}
\end{figure*}

\section{Dataset Construction}
\label{sec:dataset_cons}

\subsection{Pipeline of Data Generation}
Inspired by MemSim~\cite{zhang2024memsim}, we expand the dataset for memory evaluation based on this framework.
Building upon the question types of factual memory included in Memsim, we extend the evaluation of the ability for memory knowledge updating and extracting information from the assistant's response in single or multi-session. In addition, we incorporate reflective memory generation methods and extend observation scenario to participation scenario. The dataset creation process is as follows and as shown in Figure~\ref{fig:generate}.

\begin{figure}[t]
  \includegraphics[width=\columnwidth]{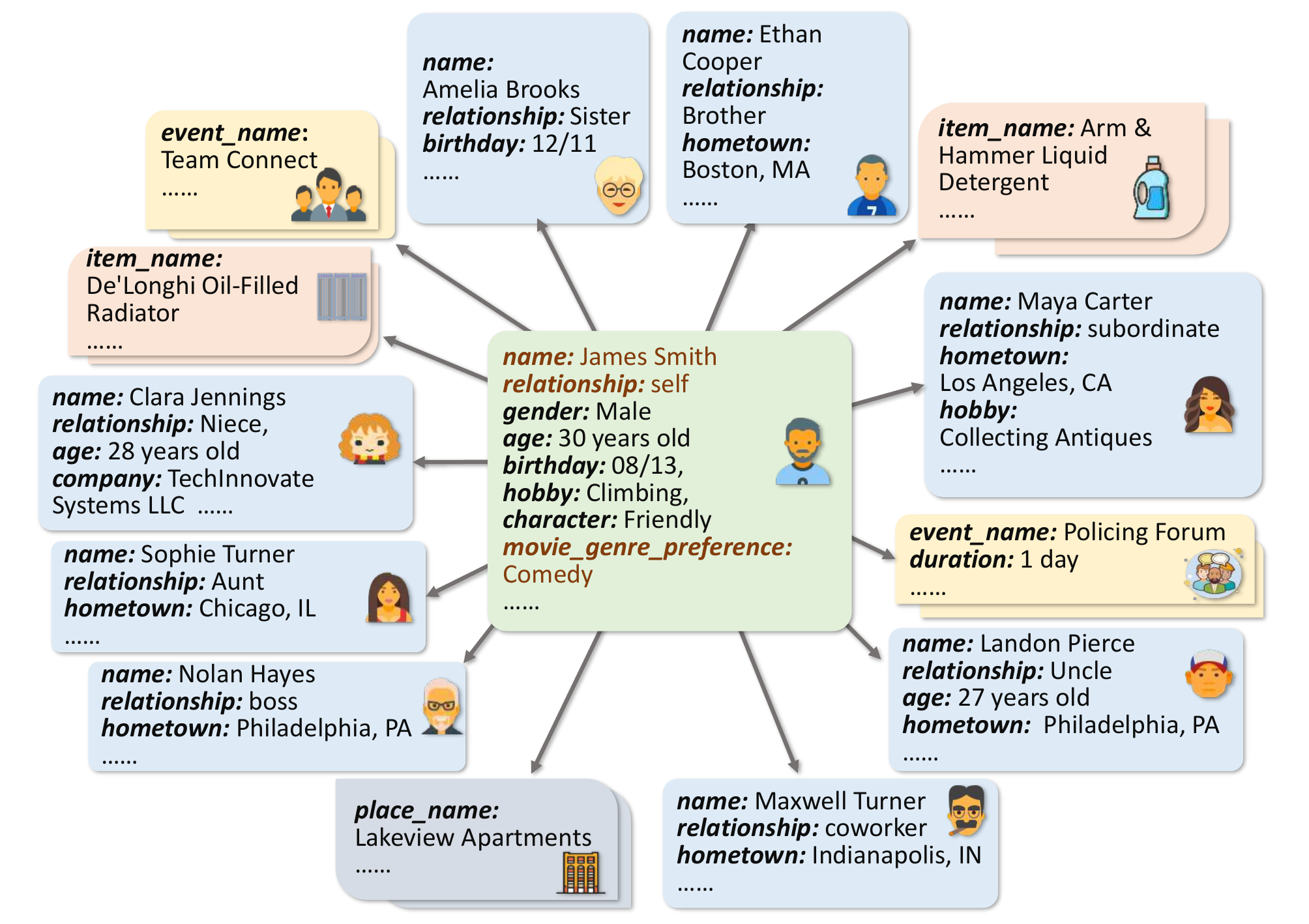}
  \vspace{-0.4cm}
  \caption{A user relation graph is composed of a user's profile and his or her associated entities including individuals, events, items and places.}
  \vspace{-0.6cm}
  \label{fig:graph}
\end{figure}

\textbf{User's Relation Graph Sampling.} Following the approach of Memsim, we create a relation graph composed of user profiles and their related entities including individuals, events, places, and items. Based on Memsim's method for sampling attributes related to factual memory, we propose a method for sampling of high-level attributes related to reflective memory. To better fit the distribution of high-level attributes in the real world, we leverage user-item relationship pairs and relevant ratings from three recommendation datasets, including MovieLens~\citep{harper2015movielens}, Food~\citep{majumder2019generating}, and Goodreads~\citep{DBLP:conf/recsys/WanM18, DBLP:conf/acl/WanMNM19}. We extract each user's high-level preferences in each recommendation dataset by identifying the most frequent category of items with which he or she likes or rates positively. If there is no category information, we utilize LLMs(GPT-4o-mini) to summarize the high-level preferences corresponding to these positive relation items. We assign high-level preference attributes using either random matching or matching based on identical attributes, and then we obtain the user's relation graph shown in Figure~\ref{fig:graph}. At the same time, we construct three one-to-many mappings between high-level preferences and low-level factual attributes with LLMs or the item-category relationships from the recommendation datasets.

\textbf{Memory Dataset Construction.} Memsim provides a data creation process for the observation scenario. We expand it to the participation scenario using the self-dialogue method. When selecting high-level preference attributes, we should choose multiple low-level preference attributes from the mapping dictionary and use these low-level preference attributes to generate the corresponding evidence dialogues. For example, a user might say, "I like the movie \textit{Star Wars}". To ensure fluency of the conversation, the specific discussions about the low-level preferences, such as the discussion about content of the movie, are inserted between the key dialogues to form a complete conversation. Finally, the remaining relevant attributes are used to generate multi-turn dialogues. The evidence dialogues will be inserted into them to form a complete session. We introduce a time-based session division approach, where the timestamp within a session is assigned continuously to each turn dialogue, typically with short intervals, such as one minute. The timestamp across different sessions maintains a sequential order, but the time gap between two adjacent sessions is typically longer, such as one day.

\begin{figure*}[t]
  \includegraphics[width=\textwidth]{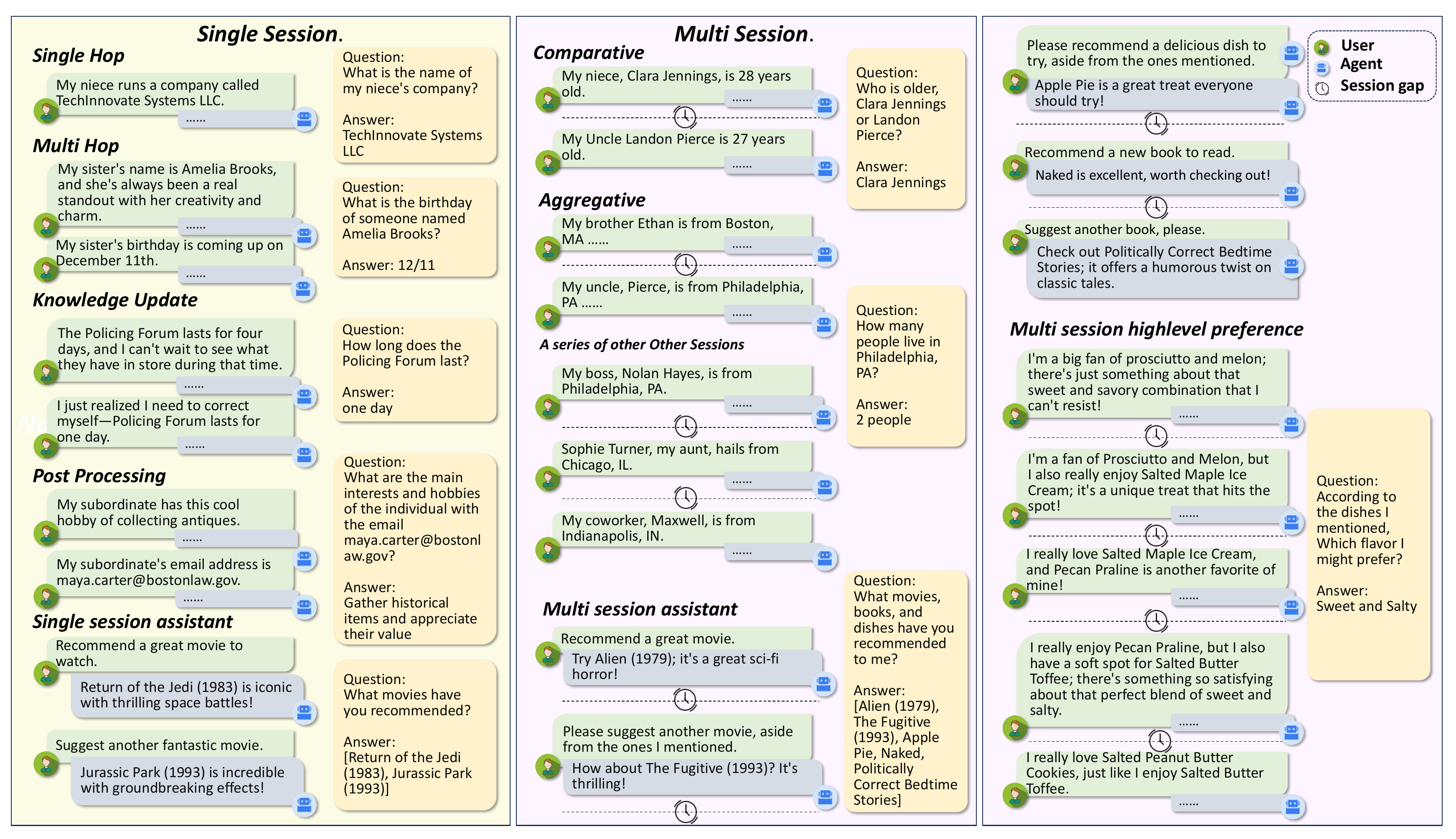}
  \vspace{-0.6cm}
  \caption{An overview of part categories of data used to test different abilities.}
  \label{fig:overview}
\end{figure*}

\begin{figure*}[h!]
    \centering
    \begin{subfigure}{0.32\linewidth}
    \vspace{-2\baselineskip}
        \centering
        \includegraphics[width=\linewidth]{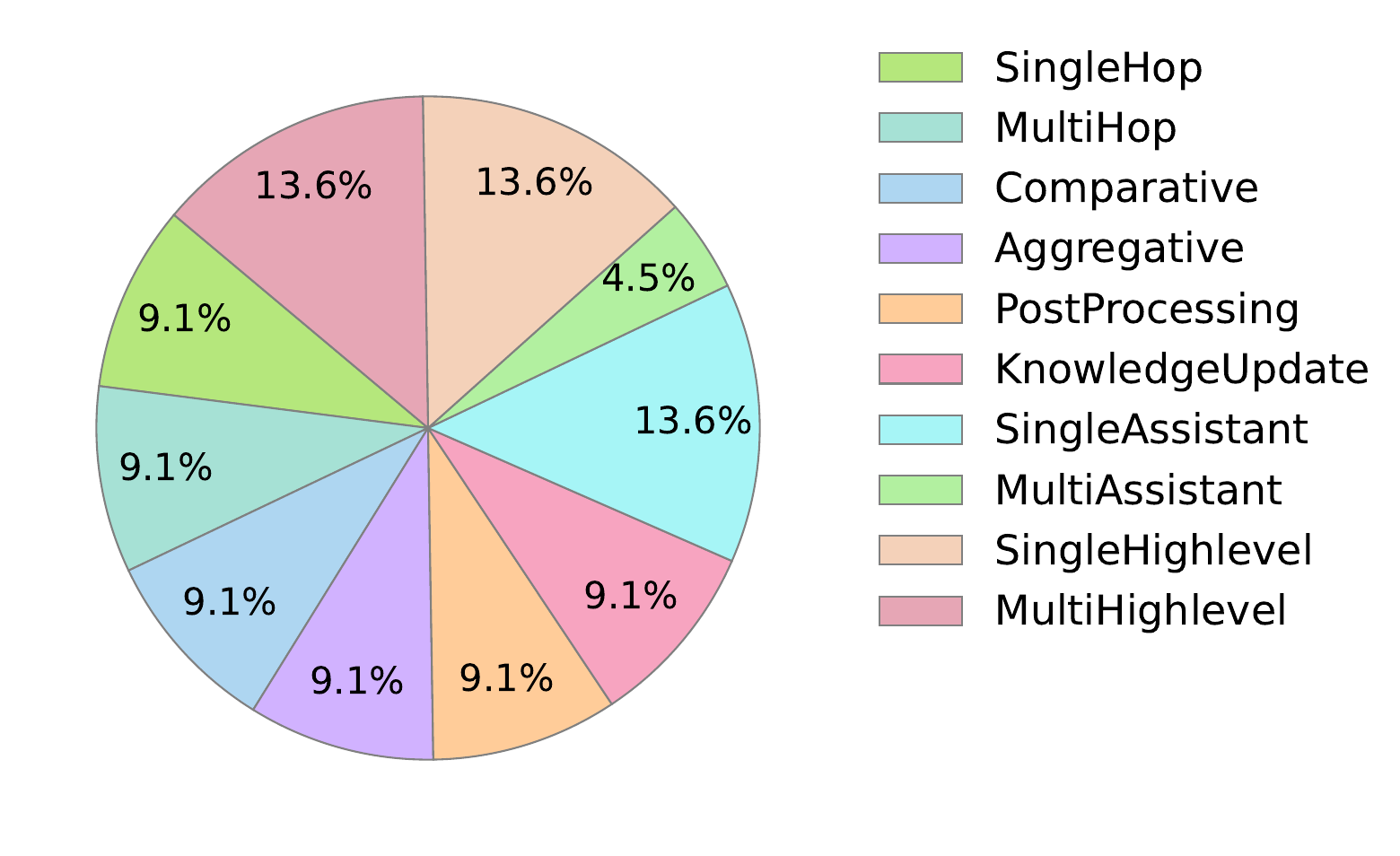}
        \subcaption{The distribution of different types of questions in the participation dataset.}
        \label{fig:sub1}
    \end{subfigure}
    \hfill
    \begin{subfigure}{0.32\linewidth}
    \vspace{-2\baselineskip}
        \centering
        \includegraphics[width=\linewidth]{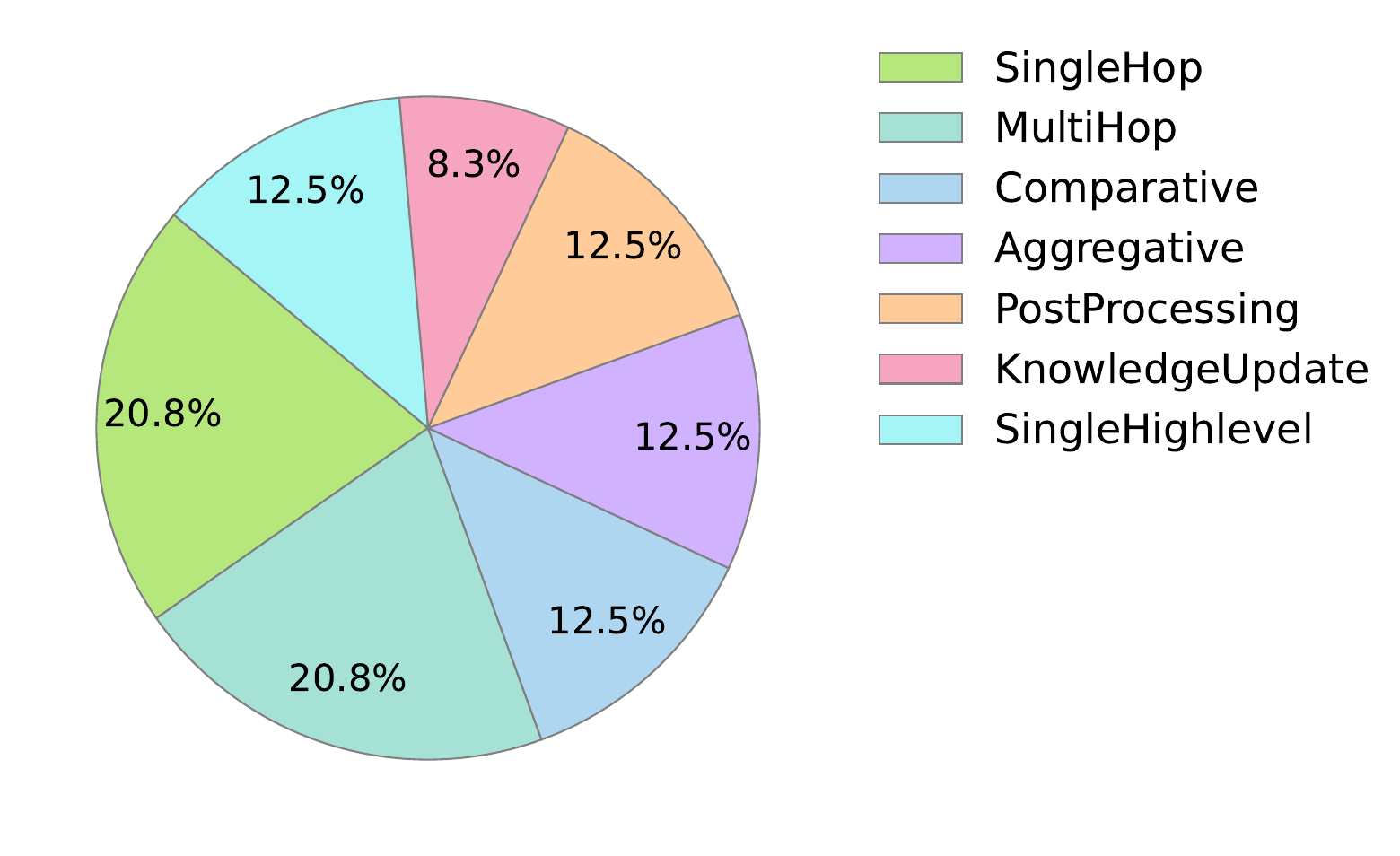}
        \subcaption{The distribution of different types of questions in the participation dataset.}
        \label{fig:sub2}
    \end{subfigure}
    \begin{tikzpicture}
    \draw[dashed] (0,0) -- (0,4); 
    \end{tikzpicture}
    \begin{subfigure}{0.32\linewidth}
    \vspace{-2\baselineskip}
        \centering
        \includegraphics[width=\linewidth]{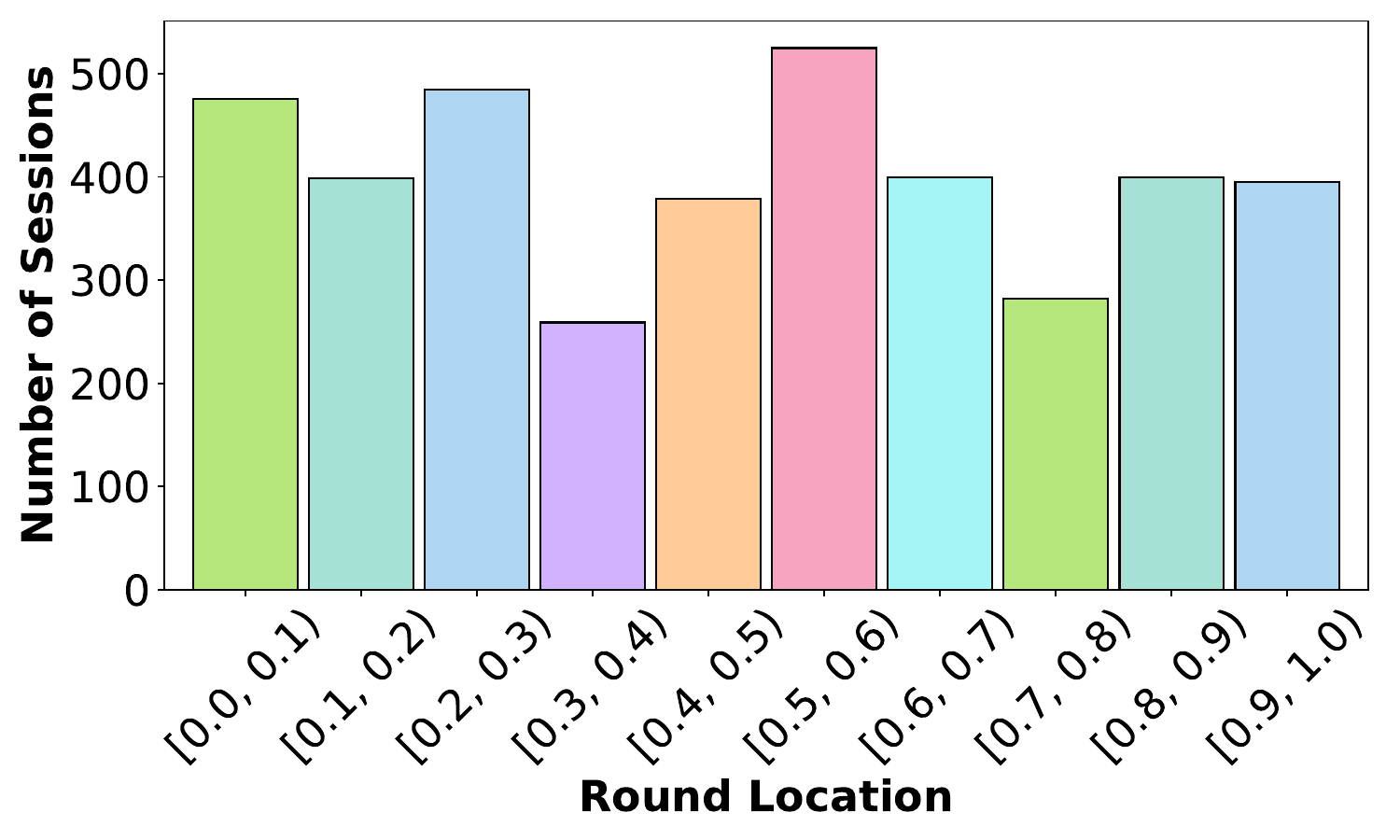}
        \subcaption{The distribution of the number of different round locations of key evidence dialogue turn within a session.}
        \label{fig:sub3}
    \end{subfigure}
    \caption{The details of the category distribution and answer distribution in the dataset.}
    \vspace{-0.5cm}
    \label{fig:three}
\end{figure*}

\subsection{Multi-scenario Memory}

The interactive scenarios of the agent can be categorized into two types, including the participation scenario and the observation scenario. In the participation scenario, the agent interacts with the user, while in the observation scenario, the agent serves only as an observer, recording user-inputted messages. In the participation scenario, other modules of the agent, such as the reasoning module, will affect the memory module. However, in the observation scenario, the agent does not perform actions and thus does not influence memory. These two scenarios cannot be considered the same. Therefore, we provide the following two types of datasets:

\textbf{Participation Memory Scenario}. The participation memory scenario is represented by the dialogue between the user and the agent, which is the agent's typical usage scenario. To eliminate the influence of other modules of the agent, we predefine the agent's responses to the user's expressions. In the user-agent dialogue interaction, the agent's memory not only needs to remember the message expressed by the user but also needs to store the message of the agent's responses, such as the agent's reply when the user requests a recommendation. In our dataset, the data for participation scenario dataset consists of sessions composed of many turns in dialogues.

\textbf{Observation Memory Scenario}. The observation memory scenario is represented by the flow of message input from the user to the agent. In this process, the agent passively receives the user's message flow over time and does not interact with the user. This scenario focuses on the agent's role as an observer, where the agent only needs to remember the message expressed by the user without taking any action. In our dataset, the data for observation scenario dataset consists of message lists composed of many messages. 

\subsection{Multi-level Memory}
In the daily usage of LLM-based agents, we expect it to have factual memory capabilities, while also hoping that its memory mechanism can summarize and reason at higher levels to generate reflective memory. Reflective memory enables the agent to gain a more comprehensive understanding of the user, thereby improving the satisfaction of subsequent interactions.
From this perspective, we divide the types of memory data and questions in our dataset into two categories:

\textbf{Factual Memory.} It refers to the specific factual attributes of the users or the entities associated with them, such as their relative's age or occupation, the time details of events and so on. This information will be expressed in daily dialogues between users and agents. Asking questions about these attributes can test various memory abilities of the agent. For example, in dialogues, the user may not directly express the time of an event but might use indirect references, such as "next Monday", we can evaluate the agent's ability to extract information and instantly convert time-related information by asking it the exact time of the day of the month of the event. In addition, we can also evaluate its ability to update knowledge based on different expressions of the same attribute over time. Furthermore, by designing questions that require the integration of multiple entities' attributes for answers, we evaluate the agent's memory capability in terms of its memory reasoning abilities in both single-session and multi-session contexts. These question examples are shown in Figure~\ref{fig:overview}.

\textbf{Reflective Memory.} Reflective memory refers to the extraction and summarization of high-level preferences based on the user's expression of low-level preferences, including some factual attributes in the dialogue. For example, the user's taste preferences are inferred from his expressions of liking for different dishes. To enhance the credibility of the answers, our memory content is reinforced through multiple expressions of different factual preferences or attributes to strengthen the understanding of the agent. We can evaluate the agent's memory mechanism's ability to extract and summarize preferences at different levels.

\subsection{Multi-metric Evaluation}
In order to more comprehensively assess the memory mechanism of the agent, we employ a total of four evaluation metrics as follows.

\textbf{Memory Accuracy.} To avoid misjudgments caused by the agent’s flexible expression of answers, in our evaluation dataset, all questions are set as multiple-choice questions. After the agent completes the memory process, both the questions and options will be submitted to the agent. The accuracy score of the memory is calculated by comparing the agent’s choice with the true choice.

\textbf{Memory Recall.} For retrieval-based memory mechanisms, the accuracy of retrieval is also an important metric that needs to be measured. It not only reflects the agent’s ability to effectively store and organize memory content, but also indicates the agent’s efficient use of memory when answering questions. In the process of creating dialogues, we first generate key evidence dialogues for answering questions, which enables the measurement of retrieval accuracy.

\textbf{Memory Capacity.} We consider that the agent's memory mechanism might have a capacity limit, which is reflected in a sharp decline in accuracy when the amount of memory content reaches a certain point. This critical threshold represents the capacity of the memory. This phenomenon might not exist, because, for example, when evaluating the retrieval-based memory mechanisms, their accuracy depends on the effectiveness of retrieval.

\textbf{Memory Efficiency.} Regarding the design of the agent’s memory mechanism, we need to focus not only on the accuracy and completeness of the memory but also on efficiency. Some memory mechanisms may result in excessively high processing time costs for the agent, which could be unacceptable in practical applications.

\subsection{Dataset Statistics}

\begin{table}[t]  
  \centering 
\caption{\revise{The statistics of our dataset. TPT indicates the average number of tokens per trajectory. PS indicates Participation Scenario. OS indicates Observation Scenario. RM indicates Reflective Memory. FM indicates Factual Memory}}
\vspace{-0.2cm}
  \resizebox{\columnwidth}{!}{
  \begin{tabular}{ccccc}
\hline
\hline
Data Type & \# Session & \# Question & \# Trajectory & TPT \bigstrut\\
\hline
PS-RM & 3.5k  & 3.5k  & 3.5k  & 2,195 \bigstrut[t]\\
PS-FM & 51k   & 39k & 8k   & 10,285 \\
OS-RM & 2k  & 2k  & 2k  & 745 \\
OS-FM & 8.5k  & 8.5k  & 8.5k  & 617 \bigstrut[b]\\
\hline
\hline
\end{tabular}%

  } 
  \vspace{-0.6cm}
  \label{tab:statistics}  
\end{table}

The dataset consists of two parts: (1) 500 graphs composed of user profiles and profiles of entities associated with users, and (2) multiple dialogues between users and assistants, multiple users' messages, and corresponding questions. The quantity is shown in Table~\ref{tab:statistics}. In order to better simulate the distribution of the location of answer in real-world conversation, the key evidence rounds in a session are almost evenly distributed in each round in a session. As shown in Figure~\ref{fig:three}, we can see the quantity distribution of different categories and the quantity distribution of key evidence rounds in the session.

\section{Benchmark}
\label{sec:benchmark}
In this section, we create a benchmark based on our dataset to evaluate the memory capabilities of LLM-based personal agents. To better evaluate the upper bounds of the agent's memory mechanism capabilities, we also utilize the News dataset~\cite{dataguy_gordon_amoako_2022} to generate a large amount of dialogues and messages serving as noise memory content that is irrelevant to the questions. We ensures that the content of noise data does not contain factual conflicts with memory messages, or dialogues in our evaluation dataset. It also allows us to control the difficulty of the evaluation by adjusting the proportion of noise data.

\begin{table*}[htbp]
  \caption{The results of different memory mechanisms on factual memory dataset. The read time (RT) and write time (WT) are presented in seconds per operation.}
  \vspace{-0.2cm}
  \label{tab:Low}
\resizebox{\textwidth}{!}{
  \centering
  \begin{tabular}{ccccccccc}
\hline
\hline
\multirow{2}[4]{*}{Method} & \multicolumn{2}{c}{Participation-Accuracy} & \multicolumn{2}{c}{Paticipation-Efficiency} & \multicolumn{2}{c}{Observation-Accuracy} & \multicolumn{2}{c}{Observation-Efficiency} \bigstrut\\
\cline{2-9}      & 10k   & 100k  & RT    & WT    & 1k    & 100k  & RT    & WT \bigstrut\\
\hline
FullMemory & 0.647 & 0.489 & 0.001 & <0.001 & 0.786 & 0.631 & <0.001 & <0.001 \bigstrut[t]\\
RecentMemory & 0.639 & 0.422 & 0.001 & <0.001 & 0.8   & 0.512 & <0.001 & <0.001 \\
RetrievalMemmory & 0.692 & 0.833 & 0.041 & 0.058 & 0.883 & 0.933 & 0.024 & 0.026 \\
GenerativeAgent & 0.478 & 0.455 & 0.045 & 6.116 & 0.779 & 0.476 & 0.031 & 6.239 \\
MemoryBank & 0.442 & 0.456 & 0.035 & 8.047 & 0.721 & 0.488 & 0.037 & 18.243 \\
MemGPT & 0.455 & 0.411 & 4.549 & 0.106 & 0.789 & 0.488 & 1.541 & 2.480 \\
SCMemory & 0.355 & 0.444 & 1.531 & 2.276 & 0.529 & 0.429 & 0.085 & 0.535 \bigstrut[b]\\
\hline
\multirow{2}[4]{*}{Method} & \multicolumn{4}{c}{Participation-Recall@10} & \multicolumn{4}{c}{Observation-Recall@10} \bigstrut\\
\cline{2-9}      & \multicolumn{2}{c}{10k} & \multicolumn{2}{c}{100k} & \multicolumn{2}{c}{10k} & \multicolumn{2}{c}{100k} \bigstrut\\
\hline
RetrievalMemmory & \multicolumn{2}{c}{0.776} & \multicolumn{2}{c}{0.749} & \multicolumn{2}{c}{0.847} & \multicolumn{2}{c}{0.769} \bigstrut\\
\hline
\hline
\end{tabular}%

}
\vspace{-0.1cm}
\end{table*}%

\begin{table*}[htbp]
  \caption{The results of different mechanisms on reflective memory dataset. The read time (RT) and write time (WT) are presented in seconds per operation.}
  \vspace{-0.2cm}
  \label{tab:high}%
\resizebox{\textwidth}{!}{
  \begin{tabular}{ccccccccc}
\hline
\hline
\multirow{2}[4]{*}{Method} & \multicolumn{2}{c}{Participation-Accuracy} & \multicolumn{2}{c}{Paticipation-Efficiency} & \multicolumn{2}{c}{Observation-Accuracy} & \multicolumn{2}{c}{Observation-Efficiency} \bigstrut\\
\cline{2-9}      & 10k   & 100k  & RT    & WT    & 1k    & 100k  & RT    & WT \bigstrut\\
\hline
FullMemory & 0.733  & 0.533  & <0.001 & <0.001 & 0.883  & 0.333  & <0.001 & <0.001 \bigstrut[t]\\
RecentMemory & 0.700  & 0.333  & <0.001 & <0.001 & 0.867  & 0.400  & <0.001 & <0.001 \\
RetrievalMemmory & 0.692  & 0.833  & 0.036  & 0.057  & 0.883  & 0.933  & 0.026  & 0.028  \\
GenerativeAgent & 0.742  & 0.333  & 0.028  & 6.064  & 0.883  & 0.200  & 0.030  & 6.019  \\
MemoryBank & 0.692  & 0.400  & 0.033  & 15.705  & 0.900  & 0.333  & 0.032  & 12.827  \\
MemGPT & 0.733  & 0.367  & 1.042  & <0.001 & 0.883  & 0.200  & 0.921  & <0.001 \\
SCMemory & 0.542  & 0.267  & 0.036  & 0.057  & 0.783  & 0.333  & 0.025  & 0.028  \bigstrut[b]\\
\hline
\hline
\end{tabular}%
}
\vspace{-0.3cm}
\end{table*}%

\subsection{Experimental Settings}
To better align with the memory process of agents in real-world scenarios, particularly the flow of time, we simulate the interaction process between the user and the agent to input the content that needs to be remembered. At time $t$, we input the user's statement from the $t$-th round, while content from the previous $t-1$ round and earlier can only be recalled through memory. In the participation memory scenario, at the $t$-th round, the agent not only needs to remember the user's messages but also needs to remember the response it has generated, which is predefined by us. Meanwhile, in the observation scenario, the agent only needs to remember the user's messages.

To set different levels of difficulty, we utilize the noise dataset to randomly insert some noise sessions into the adjacent sessions. By controlling the proportion, we create a dataset with an average length of over 100k tokens for each individual test.
Due to the large number of datasets we have created and the complexity of the agent's memory mechanism design, we perform uniform sampling on each subset of the two different sized datasets as the final tests in this paper. In the dataset of ordinary size, we extract 120 reflective memory and 360 factual memory data in participation test data (each session has about 10K tokens), as well as 60 reflective memory data and 280 factual memory data in observation test data (each message list has about 1K tokens), formulating as Sub-dataset 1. For the 100k dataset, we extract 30 reflective memory data and 90 factual memory data in participation test data (each session has about 100K tokens), as well as 15 reflective and 84 factual memory data in observation test data (each message list has about 10K tokens), denoted as Sub-dataset 2.
 
To eliminate the influence of other memory mechanism designs on the agent in the evaluation results, we make no modifications to the agent's action modules or other components.Based on MemEngine~\citep{zhang2025memengine}, we implement seven memory mechanisms, using Qwen2.5-7B as the base model for the agent applications on our benchmark, including FullMemory, RetrievalMemory, RecentMemory, GenerativeAgent~\cite{park2023generative}, MemoryBank~\cite{zhong2024memorybank}, MemGPT~\cite{packer2023memgpt}, and Self-Controlled Memory (SCMemory)~\cite{wang2023enhancing}. In our experiments, all methods that involve retrieval use the multilingual-e5-small~\cite{wang2024multilingual} for retrieval.

\subsection{Evaluations on Factual Memory}

The test results for factual memory are shown in Table~\ref{tab:Low}. FullMemory, RetrievalMemory, and RecentMemory perform better than other memory mechanisms on Sub-dataset 1. However, on Sub-dataset 2, FullMemory and RecentMemory exhibit a certain degree of decline, as the target message may fall outside the memory window. Due to the smaller window size of RecentMemory, the decline is more obvious. Other designed memory mechanisms did not show significantly superior performance in our evaluation, which might be due to flaws in these memory mechanisms. Additionally, it is important to note the time consumed by these memory mechanisms when reading and writing each round of message, especially MemGPT, which takes a longer time to read information, and MemoryBank, which takes longer to write information.

\begin{figure}[t]
    \centering
    \begin{subfigure}[b]{0.48\linewidth}
        \includegraphics[width=\linewidth]{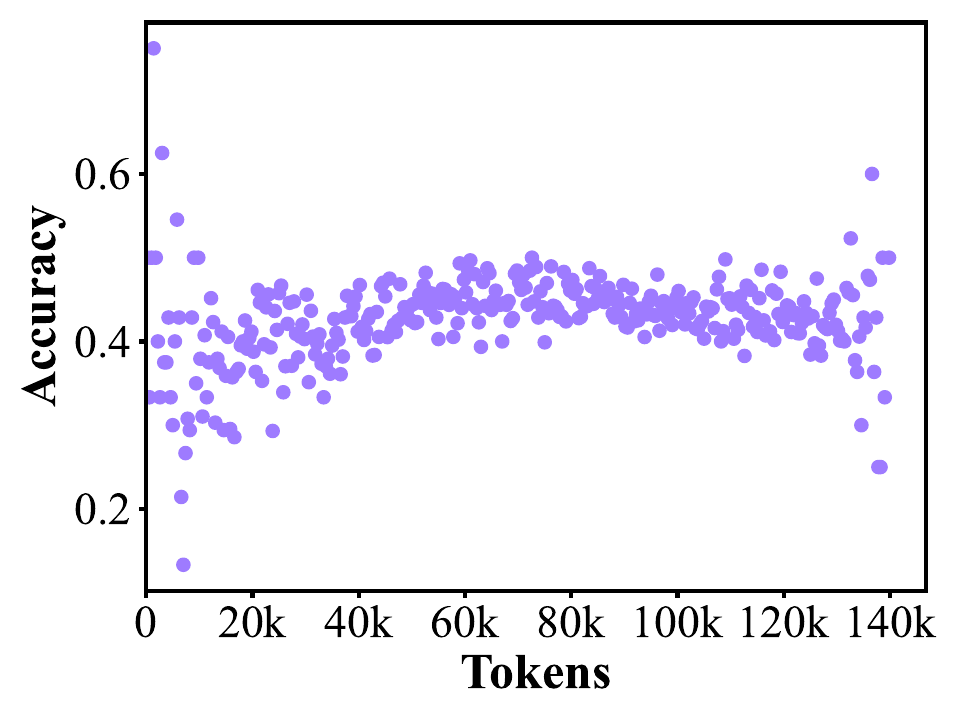}
    \end{subfigure}
    \hfil
    \begin{subfigure}[b]{0.48\linewidth}
        \includegraphics[width=\linewidth]{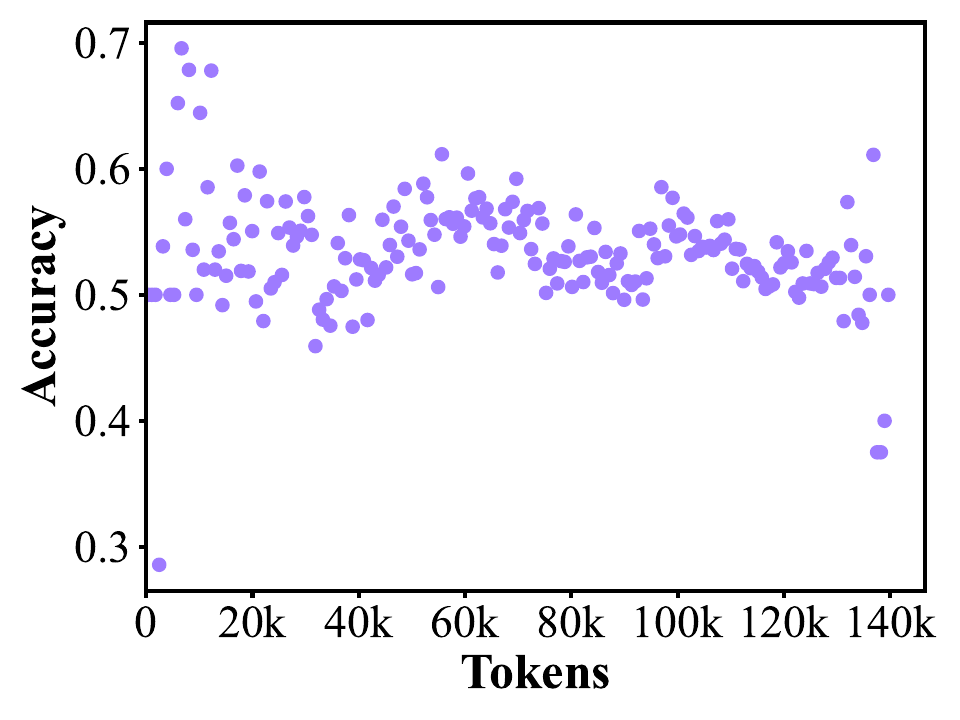}
    \end{subfigure}
    \begin{subfigure}[b]{0.48\linewidth}
        \includegraphics[width=\linewidth]{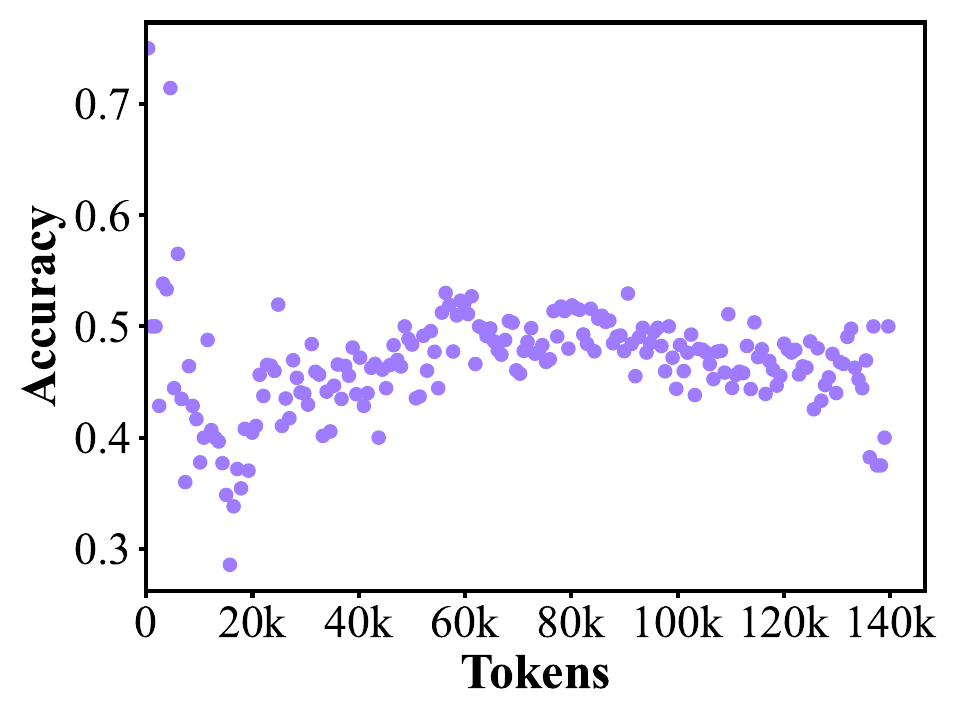}
    \end{subfigure}
    \hfil
    \begin{subfigure}[b]{0.48\linewidth}
        \includegraphics[width=\linewidth]{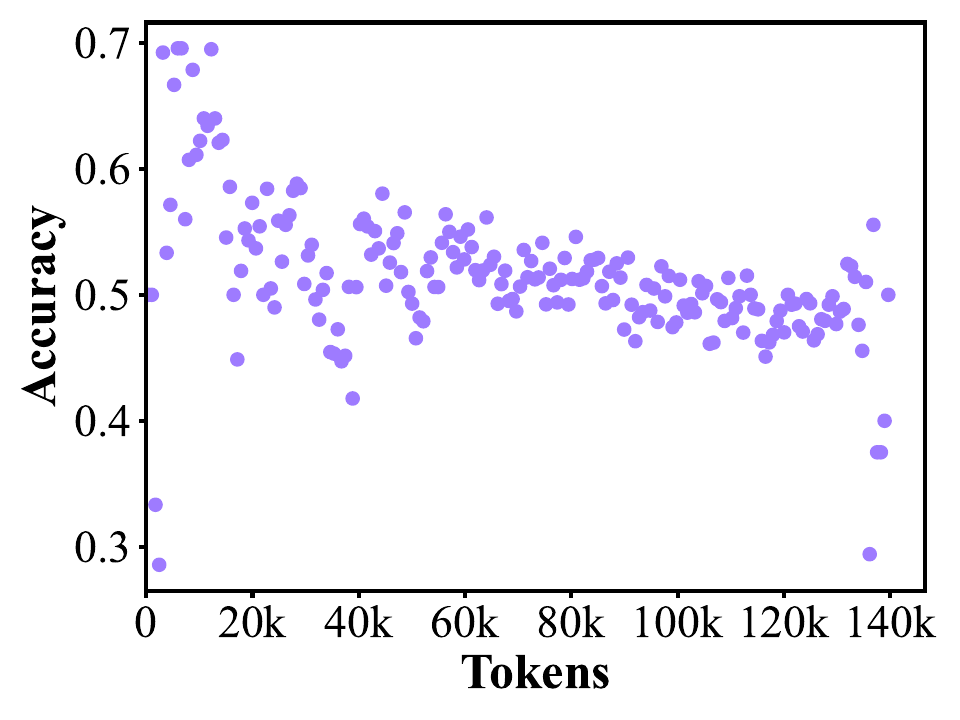}
    \end{subfigure}
    \vspace{-0.2cm}
    \caption{The accuracy of SCMemory(top-left), MemGPT(top-right), GenerativeAgent(bottom-left) and RecentMemory(bottom-right) as the memory token increases.}
    \vspace{-0.6cm}
    \label{fig:capacity}
\end{figure}

The previous evaluation works are not focusing on the design of agent memory mechanisms and solely provided factual memory datasets, so it could not adequately discuss the agent's ability to summarize reflective memory. In the following part, we evaluate on the reflective memory.

\begin{table*}[htbp]
  \centering
  \caption{The results of memory mechanisms with different LLMs on our sub-dataset 1. The read time (RT) and write time (WT) are presented in seconds per operation. For the reflective memory, P-Accuracy means the accuracy under the participant scenario, and O-Accuracy refers to the accuracy under the observation scenario.}
  \label{tab:comparison}%
  \vspace{-0.1cm}
  \resizebox{\textwidth}{!}{
\begin{tabular}{c>{\centering\arraybackslash}p{2.0cm}cc>{\centering\arraybackslash}p{2.0cm}cc>{\centering\arraybackslash}p{2.0cm}>{\centering\arraybackslash}p{2.0cm}}
\hline
\hline
\multirow{2}[4]{*}{Method} & \multicolumn{3}{c}{Factual-Participation} & \multicolumn{3}{c}{Factual-Observation} & \multicolumn{2}{c}{Reflective Memory} \bigstrut\\
\cline{2-9}      & Accuracy & RT    & WT    & Accuracy & RT    & WT    & P-Accuracy & O-Accuracy \bigstrut\\
\hline
\multicolumn{9}{c}{Qwen2.5-7B-Instruct} \bigstrut\\
\hline
FullMemory & 0.647  & 0.001  & <0.001 & 0.786  & <0.001 & <0.001 & 0.733  & 0.883  \bigstrut[t]\\
RecentMemory & 0.639  & 0.001  & <0.001 & 0.800  & <0.001 & <0.001 & 0.700  & 0.867  \\
RetrievalMemmory & 0.692  & 0.041  & 0.058  & 0.883  & 0.024  & 0.026  & 0.692  & 0.883  \\
GenerativeAgent & 0.478  & 0.045  & 6.116  & 0.779  & 0.031  & 6.239  & 0.742  & 0.883  \bigstrut[b]\\
\hline
\multicolumn{9}{c}{GPT-4o-mini} \bigstrut\\
\hline
FullMemory & 0.736  & 0.001  & <0.001 & 0.864  & <0.001 & <0.001  & 0.783  & 0.883  \bigstrut[t]\\
RecentMemory & 0.697  & 0.001  & <0.001 & 0.864  & <0.001 & <0.001  & 0.758  & 0.900  \\
RetrievalMemmory & 0.633  & 0.003  & 0.031  & 0.857  & 0.023  & 0.023  & 0.767  & 0.900  \\
GenerativeAgent & 0.592  & 0.107  & 0.970  & 0.846  & 0.030  & 0.998  & 0.758  & 0.900  \bigstrut[b]\\
\hline
\multicolumn{9}{c}{Meta-Llama-3.1-8B-Instruct} \bigstrut\\
\hline
FullMemory & 0.519  & 0.001  & <0.001 & 0.779  & <0.001 & <0.001  & 0.708  & 0.817  \bigstrut[t]\\
RecentMemory & 0.461  & 0.001  & <0.001 & 0.779  & <0.001 & <0.001  & 0.683  & 0.850  \\
RetrievalMemmory & 0.500  & 0.050  & 0.062  & 0.700  & 0.044  & 0.049  & 0.733  & 0.833  \\
GenerativeAgent & 0.430  & 0.036  & 6.551  & 0.725  & 0.065  & 12.322  & 0.725  & 0.850  \bigstrut[b]\\
\hline
\multicolumn{9}{c}{glm-4-9b-chat} \bigstrut\\
\hline
FullMemory & 0.475  & 0.001  & <0.001 & 0.775  & <0.001 & <0.001  & 0.658  & 0.850  \bigstrut[t]\\
RecentMemory & 0.539  & 0.001  & <0.001 & 0.746  & <0.001 & <0.001  & 0.708  & 0.850  \\
RetrievalMemmory & 0.483  & 0.032  & 0.037  & 0.739  & 0.025  & 0.025  & 0.742  & 0.800  \\
GenerativeAgent & 0.439  & 0.050  & 0.165  & 0.718  & 0.030  & 0.111  & 0.675  & 0.900  \bigstrut[b]\\
\hline
\hline
\end{tabular}%
}
\vspace{-0.3cm}
\end{table*}%

\subsection{Evaluations on Reflective Memory}
The test results for reflective memory are shown in Table~\ref{tab:high}. It can be observed that GenerativeAgent, MemGPT, and MemoryBank performed very well on Sub-dataset 1, but their performance significantly declines on Sub-dataset 2. Only the retrieval-based RetrievalMemory achieved the remaining good results. It is likely due to the limited context window of the models or the incorporation of forgetting mechanisms in these memory systems, which leads to the loss of important memories. However, these findings still suggest that well-designed memory mechanisms are capable of effectively capturing reflective memory. How to maintain this ability after prolonged interactions may pose a challenging research problem.

\subsection{Evaluations on Memory Capacity}

To explore the capacity of the agent's memory mechanism, we test the answering accuracy of each round after the key evidence turns on the observation scenario in Sub-dataset 2(100k). In order to observe the changes in the accuracy of MemGPT and Self-Controlled Memory with the number of tokens increases, we drew Figure~\ref{fig:capacity}. From the results, we can observe that both memory mechanisms exhibit a sharp decline, which may be due to the upper limit of memory performance retention capacity for these memory mechanisms in Qwen2.5-7B-Instruct~\cite{qwen2, qwen2.5}.

\subsection{Comparison of Different Inference Models} 
In practical applications of agents, different models may be selected for different memory mechanisms. Therefore, we evaluate the performance of several common models across various memory mechanisms. Specifically, we selected Qwen2.5-7B-Instruct, gpt-4o-mini, Meta-Llama-3.1-8B-Instruct and glm-4b-chat~\cite{glm2024chatglm} for evaluation. The results are shown in Table~\ref{tab:comparison}. Under the same context window length, the choice of base model significantly affects the agents' performence. In most cases, GPT-4o-mini performs as the best model compared to others. Although the factual memory capability of Meta-Llama-3.1-8B-Instruct is notably inferior to that of other models, its reflective memory ability is still relatively good. Interestingly, for GenerativeAgent, choosing GPT-4o-mini as the base model results in a significantly higher time consumption compared to other models in our experiments. However, in most cases, the time consumption differences between the three models are not substantial.

\section{Conclusion}
\label{sec:conclusion}
This paper provides a more comprehensive and scalable dataset for evaluating LLM-based agent's memory mechanisms. It includes a dataset with multi-scenarios (both participation and observation), and multi-level memory content include reflective memory and factual memory. Based on this dataset, we constructed a time-aware evaluation framework that simulates the daily interactions between users and agents with multi-metric include accuracy, recall, capacity and temporal efficiency. We evaluate the performance of seven common memory mechanisms in agents on our benchmark.

\section*{Limitations}
The dataset proposed in this paper consists of a graph formed by the profiles of users and relevant entities, enabling further exploration of the agent's memory mechanism. Our evaluation method is limited by an assessment of memory for structured data. However, by comparing the construction of relevant entity profiles or the capture of specific attribute information in the agent’s memory during user-agent interactions, we can investigate the agent's ability to structure memory. In addition, there are still many areas to explore in reflective memory, such as users' emotional memory.

\section*{Ethics Statement}
The data used in this article to construct the dataset includes data from publicly available, authorized datasets.
All publicly available data are used in accordance with their respective licenses for research purposes. The LLM-generated content may pose risks, including the potential for unintended biases or harmful output. Although we have taken steps to minimize these risks, we encourage users to apply the dataset responsibly to avoid ethical risks.

\section*{Acknowledgments}
This work is supported in part by National Natural Science Foundation of China (No. 62422215 and No. 62472427), Major Innovation \& Planning Interdisciplinary Platform for the “DoubleFirst Class” Initiative, Renmin University of China, Public Computing Cloud, Renmin University of China,  fund for building world-class universities (disciplines) of Renmin University of China.
This work is also sponsored by Huawei Innovation Research Programs. We gratefully acknowledge the support from Mindspore\footnote{\url{https://www.mindspore.cn}}, CANN(Compute Architecture for Neural Networks) and Ascend AI Processor used for this research.

\bibliography{custom}
\clearpage

\appendix

\section{Case Studies}
\label{sec:appendix}
\subsection{User Relation Graph Example}

In this section, we present examples of the components of our dataset, including the user graph and test cases. For the user graph, we show profile examples of the user itself, related individuals, events, items, and places.

\textbf{UserProfile:} "gender": "Male", "relationship": "self", "name": "James Smith", "age": "30 years old", "height": "164 cm", "birthday": "08.13", "hometown": "San Francisco, CA", "work location": "Boston, MA", "education": "Associate Degree", "occupation": "Police Officer", "position": "Community Policing Officer", "company name": "Boston Law Enforcement Agency", "hobby": "Climbing", "character": "Friendly", "contact number": "4150430511", "email address": "james.smith@bostonlawenforcement.gov", "ssn": "914610199408130162", "passport number": "PUP4822676", "bank account": "6222022865544246", "driver license": "914EAPRDV5F", "highlevel preference": ["movie genre preference": ("Comedy", "Romance", "Action", "Drama", "Thriller"), "taste preference": "Umami and Sweet", "book preference": "Humor"].

\textbf{RelativeRoleProfile.}"gender":"Male", "relationship":"Brother","name": "Ethan Cooper","age": "28 years old","height": "165cm","birthday": "01/20","hometown": "Boston, MA", "work location": "Los Angeles, CA", "education": "Associate Degree", "occupation": "Electrician", "position": "Electrical Maintenance Technician", "company name": "SparkLight Electric Services", "hobby": "Running", "character": "Thoughtful", "contact number": "20103787263","email address": "ethan.cooper@sparklightelectric.com"

\textbf{ColleagueRoleProfile.} "gender": "Male", "relationship": "boss", "name": "Nolan Hayes", "age": "39 years old","height": "170cm","birthday": "03/24","hometown": "Philadelphia, PA","work location": "Boston, MA","education": "Associate Degree","occupation": "Police Officer","position": "Police Sergeant","company name": "Boston Law Enforcement Agency", "hobby": "Attending Concerts","character": "Empathetic", "contact number": "30503926075","email address": "nolan.hayes@bostonlawagency.gov"

\textbf{WorkEventProfile.} "event type": "Company Team Building", "main content": "Community engagement workshop for team bonding.", "location": "Los Angeles, CA", "time": "the week after next Sat 9:00 AM", "event name": "Team Connect", "scale": "one hundred people", "duration": "eight day"

\textbf{RestEventProfile.}"event type": "Community Fair", "main content": "Join us at the Community Fair for climbing challenges, equipment demos, and safety workshops! Engage with fellow enthusiasts, explore local climbing spots, and enjoy inspiring talks from seasoned climbers. Perfect for all skill levels and climbing enthusiasts!", "location": "Miami, FL", "time": "2024-10-12 19:00", "event name": "Climb Fest", "scale": "nine hundred people", duration": "three day"

\textbf{ItemProfile:} "relationship": "Own", "item type": "Laundry Detergent", "item name": "Arm \& Hammer Liquid Detergent", "item review": "As a police officer, I’m always on the go, and I need products that can keep up with my busy lifestyle. I've been using Arm \& Hammer Liquid Detergent for a while now, and I have to say, it’s been a game changer for me. Not only does it tackle tough stains from my uniforms and gear with ease, but it also leaves my clothes smelling fresh and clean. The added baking soda really helps to neutralize odors, which is a must when you're working in various environments. Plus, I appreciate that it’s available in eco-friendly options, making it easier to care for the planet while looking after my laundry. Definitely a solid choice for anyone looking for effective and reliable detergent!"

\textbf{PlaceProfile.}
Place example: "relationship": "Visited","place type": "Mall", "place name": "The Grove", "place review": "I recently visited The Grove and I have to say, it was a really refreshing experience. The vibe there is incredibly friendly and welcoming, just like the community I strive to serve as a police officer. The shops and restaurants offer a great variety, and I especially enjoyed grabbing a bite at one of the local eateries. The layout is easy to navigate, making it a perfect spot to relax and enjoy some fresh air.As someone who loves climbing, I appreciated the green spaces where you can unwind and enjoy nature. It's a fantastic place to spend time with family or friends. The only downside I found was that it got a bit crowded during peak hours, but that's to be expected in such a popular location. Overall, I’d highly recommend The Grove to anyone looking for a fun and friendly outing!"

\begin{table*}[htbp]
\caption{Overview of Factual Memory questions.}
\label{tab:6}
\centering
\resizebox{\textwidth}{!}{
\begin{tabular}{cl}
\hline
\hline
\textbf{Types} & \textbf{Descriptions} \\ 
\hline
Single-hop & Rely on one message to answer the question directly.\\ 
\hline
Multi-hop  & Require multiple messages to answer the question jointly. \\ 
\hline
Comparative & Compare two entities on a shared attribute with multiple messages.\\ 
\hline
Aggregative  & Aggregate messages about more than two entities on a common attribute.  \\ 
\hline
Post-processing & Involve extra reasoning steps to answer with multiple messages.\\ 
\hline
Knoewledge-updating & The basis for answering questions will be updated over time with different messages.\\ 
\hline
Single-session-assistant & Rely on a single message from the assistant to directly answer the question.\\ 
\hline
Multi-session-assistant & Require Multiple messages from the assistant to collectively answer the question.\\ 
\hline
\hline
\end{tabular}
}
\end{table*}

\begin{table*}[htbp]
\centering
\vspace{-0.3cm}
\caption{Overview of Reflective Memory questions.}
\label{tab:7}
\vspace{-0.3cm}
\resizebox{\textwidth}{!}{
\begin{tabular}{cl}
\hline
\hline
\textbf{Types} & \textbf{Descriptions} \\ 
\hline
Preference & Rely on multiple messages to actively express the user's lower-level preferences.\\ 
\hline
Emotion  & Rely on multiple consecutive messages within a specific time to express the user's emotional state. \\ 
\hline
\hline
\end{tabular}
}
\end{table*}

\subsection{FM-RM Directionary Example}

When creating the correspondence between factual memory attribute and reflective memory attribute, we simultaneously created a dictionary mapping the two. Below, we provide examples from each category of reflective memory.

\textbf{Movie.} "Action": ["Star Wars (1977)","Godfather, The (1972)","Raiders of the Lost Ark (1981)","Titanic (1997)","Empire Strikes Back, The (1980)","Boot, Das (1981)","Godfather: Part II, The (1974)","African Queen, The (1951)","Princess Bride, The (1987)","Braveheart (1995)", "Glory (1989)", "Fugitive, The (1993)","Alien (1979)","Return of the Jedi (1983)","Terminator 2: Judgment Day (1991)", "Butch Cassidy and the Sundance Kid (1969)","Aliens (1986)","Magnificent Seven, The (1954)","Terminator, The (1984)","Apollo 13 (1995)","Indiana Jones and the Last Crusade (1989)","Die Hard (1988)","Hunt for Red October, The (1990)","Good, The Bad and The Ugly, The (1966)","Blues Brothers, The (1980)","Ben-Hur (1959)", "Cyrano de Bergerac (1990)", "Star Trek: The Wrath of Khan (1982)","In the Line of Fire (1993)", "Adventures of Robin Hood, The (1938)","Jaws (1975)","Face/Off (1997)","Men in Black (1997)","Diva (1981)","Jurassic Park (1993)","Rock, The (1996)","Full Metal Jacket (1987)", "Perfect World, A (1993)","Star Trek: First Contact (1996)","Speed (1994)","Air Force One (1997)", "Crying Game, The (1992)", "True Romance (1993)","Abyss, The (1989)","Clear and Present Danger (1994)","Heat (1995)","True Lies (1994)","Get Shorty (1995)","Last of the Mohicans, The (1992)","Supercop (1992)"]

\textbf{Food.}
"Sweet": ["Candy", "Honey", "Fruit", "Maple Syrup Pancakes", "Baklava", "Chocolate Cake", "Custard", "Jelly", "Pecan Pie", "Apple Pie", "Brownies", "Banana Bread", "Donuts", "Rice Krispies"]

\textbf{Book.}"Health \& Fitness": ["What to Expect When You're Expecting (Revised Edition)",
"Make the Connection: Ten Steps to a Better Body and a Better Life",
"The South Beach Diet: The Delicious, Doctor-Designed, Foolproof Plan for Fast and Healthy Weight Loss",
"Dr. Atkins' New Diet Revolution",
"Dr. Atkins' New Diet Revolution",
"Prescription for Nutritional Healing: A Practical A-Z Reference to Drug-Free Remedies Using Vitamins, Minerals, Herbs \& Food Supplements",
"8 Weeks to Optimum Health",
"Body for Life: 12 Weeks to Mental and Physical Strength",
"Your Pregnancy: Week by Week (Your Pregnancy Series)",
"Fat Land: How Americans Became the Fattest People in the World"]

\subsection{Question Description}
In Tab~\ref{tab:6}, we provide an explanation for each type of Factual Memory questions.
In Tab~\ref{tab:7}, we provide an explanation for each type of Reflective Memory questions.

\subsection{Participation Example}

In this section, we provide more detailed examples from the participation scenarios in our dataset. To make the presentation clearer, we have only listed the key evidence dialogue rounds necessary for answering the questions and omitted any unnecessary information. 

The detailed examples are as follows:

\textbf{Single Hop}

User: My niece runs a company called TechInnovate Systems LLC.

Assistant:...

Question: What is the name of my niece's company?

Answer: TechInnovate Systems LLC

\textbf{Multi Hop}

User: My sister's name is Amelia Brooks, and she's always been a real standout with her creativity and charm.

Assistant:...

User: My sister's birthday is coming up on December 11th.

Assistant:...

Question: What is the birthday of someone named Amelia Brooks?

Answer: 12/11

\textbf{Knowledge Updating}

User:The Policing Forum lasts for four days, and I can't wait to see what they have in store during that time.
Assistant:

User: I just realized I need to correct myself—Policing Forum only lasts for one day.
Assistant:...

Question: How long does the Policing Forum last?

Answer: one day

\textbf{Post Processing}

User: My subordinate has this cool hobby of collecting antiques.

Assistant:...

User: My subordinate's email address is maya.carter@bostonlaw.gov.

Assistant:...

Question: What are the main interests and hobbies of the individual with the email address 
maya.carter@bostonlaw.gov?

Answer: Gather historical items and appreciate their value

\textbf{Single session assistant}

user: I'm looking for a great movie to watch, aside from the ones I've mentioned earlier.

assistant: I'm all about Return of the Jedi; it's such an iconic film with those thrilling space battles and unforgettable characters that make it a must-see!

user: I'm looking for a fantastic movie to watch, aside from the ones I've mentioned earlier.

assistant: I recommend the Jurassic Park from 1993; it's such an incredible film with its groundbreaking special effects and thrilling adventure that really grabs my attention!

Question: What movies have you recommended to me before?

Answer: ["Return of the Jedi (1983)","Jurassic Park (1993)"]

\textbf{Comparative}

User: My niece, Clara Jennings, is 28 years old.

Assistant: ...

User: My Uncle Landon Pierce is 27 years old.

Assistant: ...

Question: Who is older, Clara Jennings or Landon Pierce?

Answer: Clara Jennings

\textbf{Aggregative}

User: My brother Ethan Cooper hails from Boston, MA

Assistant:...

User: My sister, Amelia Brooks, is from Washington, DC.

Assistant:...

User: My niece, Clara Jennings, is from Jacksonville, FL.

Assistant:...

User: My uncle, Landon Pierce, hails from Philadelphia, PA.

Assistant:...

User: Sophie Turner, my aunt, is from Chicago, IL.

Assistant:...

User: My boss, Nolan Hayes, is from Philadelphia, PA.

Assistant:...

User: Maya Carter, who works under me, is from Los Angeles, CA.

Assistant:...

User:My coworker, Maxwell Turner, is from Indianapolis, IN.

Assistant:...

Question: How many people live in Philadelphia, PA?

Answer: 2 people

\textbf{Multi session assistant}

User: I'm looking for a great movie to watch, excluding the ones mentioned earlier.

Assistant: I definitely think Alien (1979) is a must-see; it's an intense blend of science fiction and horror that really keeps you on the edge of your seat!

User: I'm looking for a great movie to watch, aside from the ones I mentioned earlier.

Assistant: "I'm all about The Fugitive from 1993; it's a thrilling ride that really keeps you on the edge of your seat!

User: I'm looking for a delicious dish to try, aside from the ones mentioned earlier.

Assistant: I'm a big fan of Apple Pie; it's such a delightful treat that everyone should definitely give it a try!

User: I'm looking for a good book to read, excluding the ones mentioned earlier.

Assistant: I really think Naked is a fantastic read; it's definitely worth checking out!

User: I'm looking for a good book to read, aside from the ones mentioned earlier.

Assistant: I really think everyone should check out Politically Correct Bedtime Stories: Modern Tales for Our Life and Times; it offers a hilarious twist on classic tales that resonate with today’s issues.

Question: What movies, books and dishes have you recommended to me?

Answer:  ["Alien (1979)","Fugitive, The (1993)","Apple Pie","Naked","Politically Correct Bedtime Stories: Modern Tales for Our Life and Times"]

\textbf{Multi session highlevel preference}

User: I'm a big fan of prosciutto and melon; there's just something about that sweet and savory combination that I can't resist!

Assistant:...

User: I'm a fan of Prosciutto and Melon, but I also really enjoy Salted Maple Ice Cream; it's a unique treat that hits the spot!

Assistant: ...

User: I really love Salted Maple Ice Cream, and Pecan Praline is another favorite of mine!

Assistant: ...

User: I really enjoy Pecan Praline, but I also have a soft spot for Salted Butter Toffee; there's something so satisfying about that perfect blend of sweet and salty.

Assistant: ...

User: I really love Salted Peanut Butter Cookies, just like I enjoy Salted Butter Toffee.

Assistant:...

Question: According to the dishes I mentioned, Which flavor I might prefer?

Answer: Sweet and Salty

\subsection{Observation Example}

The only difference between the data for participation and the data here is the absence of responses from "assistant," so specific examples are not provided here.

\section{Detail Data Statics}
In Tab~\ref{tab:8}, we provide the detail statistics of Participation dataset. In Tab~\ref{tab:9}, we provide the detail statistics of Observation dataset.
\begin{table}[t]  
  \centering 
\caption{The detail statistics of our Participation dataset. RM indicates Reflective Memory. FM indicates Factual Memory. The types we have include ssh(sigle-hop), mh(multi-hop), comp(comparative), agg(aggregative), pp(post-processing), ku(knowledge-update), ssa(signle-session-assistant), msa(multi-session-assistant).}
\label{tab:8}
\vspace{-0.2cm}
  \resizebox{\columnwidth}{!}{
  \begin{tabular}{cccc}
\hline
\hline
Data Type & \# Session & \# Question & \# Trajectory \bigstrut\\
\hline
RM-Pr & 3.0k  & 3.0k  & 3.0k  \bigstrut[t]\\
RM-Em & 0.5k   & 0.5k & 0.5k    \\
FM-sh & 8k  & 8k  & 1k   \\
FM-mh & 8k  & 8k  & 1k  \bigstrut[b]\\
FM-comp & 8k  & 4k  & 1k   \\
FM-agg & 8k  & 1k  & 1k   \\
FM-ku & 8k  & 8k  & 1k   \\
FM-pp & 8k  & 8k  & 1k   \\
FM-ssa & 1.5k  & 1.5k  & 1.5k   \\
FM-msa & 1.5k  & 0.5k  & 0.5k   \\
\hline
\hline
\end{tabular}%
  } 
  \vspace{-0.4cm}
  \label{tab:3}  
\end{table}

\begin{table}[t]  
  \centering 
\caption{The detail statistics of our Participation dataset. RM indicates Reflective Memory. FM indicates Factual Memory. The types we have include sh(sigle-hop), mh(multi-hop), comp(comparative), agg(aggregative), pp(post-processing), ku(knowledge-update).}
\label{tab:9}
\vspace{-0.2cm}
  \resizebox{\columnwidth}{!}{
  \begin{tabular}{cccc}
\hline
\hline
Data Type & \# Session & \# Question & \# Trajectory \bigstrut\\
\hline
RM-Pr & 1.5k  & 1.5k  & 1.5k  \bigstrut[t]\\
RM-Em & 0.5k   & 0.5k & 0.5k    \\
FM-sh & 1.5k  & 1.5k  & 1.5k   \\
FM-mh & 1.5k  & 1.5k  & 1.5k  \bigstrut[b]\\
FM-comp & 1.5k  & 1.5k  & 1.5k   \\
FM-agg & 1.5k  & 1.5k  & 1.5k   \\
FM-pp & 1.5k  & 1.5k  & 1.5k   \\
FM-ku & 1k  & 1k  & 1k   \\
\hline
\hline
\end{tabular}%
  } 
  \vspace{-0.6cm}
  \label{tab:3}  
\end{table}

\section{Data Creation Prompt}

\subsection{Profile Prompt}

\textbf{Flavour Reflective Memory Attribute}
Please choose user's taste from [Tastes] according to the dishes he likes below.
[Dishes]:\{Dishes\}
[Tastes]: ["Sweet", "Sour", "Spicy", "Salty", "Umami", "Bitter", "Sweet and Salty", "Sweet and Sour", "Salty and Umami", "Sour and Spicy", "Sweet, Salty, and Spicy", "Sour and Salty", "Sour, Sweet, and Salty", "Salty, Umami, and Spicy", "Numbing and Spicy", "Creamy and Sweet", "Umami and Sweet", "Bitter and Sweet", "Astringent",  "Numbing", "Rich and Fatty", "Cool", "Warm and Spicy"]
example output:\{\{'taste': 'sweet'\}\}

\subsection{Self-dialogue Prompt}
\textbf{Role Dialogue Generation Prompt.} Please generate a \{round\_length\}-round interactive conversion between the user and assistant, with a total of \{sentence\_length\} sentences. The dialogue's main content should based on the given information about the user's \{entity\}.
Ensure that no information beyond what is provided is introduced in the dialogue. 
**Note that the user cannot ask the assistant for information because the assistant does not know the information.**
Note that the assistant is the user's personal assistant, so it should only respond passively to the user's dialogue, but it can reply with varied content. 
Please return the conversation in a JSON list format as shown in the example, ensuring that the result can be directly parsed by json.loads.
**Every json must includes both user and assistant with their words! Every json's format is \{\{"user": user's words, "assistant": assistant's words\}\}** 
\#[Information]: 
\{information\}

\#example:
[\{\{"user": "I wanted to talk to you about my cousin, Ethan Parker. He's 39 years old.", "assistant": "Certainly! Ethan is 39. Is there something specific you'd like to discuss about him?"\}\},
\{\{"user": "Well, he's actually pretty short for a guy, only 162 cm. He always jokes about it, though.", "assistant": "It sounds like Ethan has a good sense of humor about his height. That's always a great quality!"\}\},
\{\{"user": "Yes, he really does. It's one of the things that makes him so fun to be around.", "assistant": "I can imagine! Having a cousin with a lighthearted attitude must make family gatherings enjoyable."\}\},
\{\{"user": "Definitely. And since we’re so close in age, it feels like he’s more of a friend than just family.", "assistant": "That sounds wonderful. It must be special to have that kind of bond with your cousin."\}\}
\{\{"user": "It really is. We’ve shared a lot of memories growing up together.", "assistant": "Those shared memories must make your relationship even stronger. It sounds like Ethan has been a big part of your life."\}\}]

\textbf{Event Dialogue Generation Prompt.} Please generate a \{round\_length\}-round interactive conversion between the user and assistant, with a total of \{sentence\_length\} sentences. The dialogue's main content should based on the given information about the \{event\_name\}.
Ensure that no information beyond what is provided is introduced in the dialogue.
Note that the assistant is the user's personal assistant, so it should only respond passively to the user's dialogue, but it can reply with varied content. 
Note that the user cannot ask the assistant for information because the assistant does not know the information.
You can start with user saying I'm going to attend \{event\_name\}
Please return the conversation in a JSON list format as shown in the example, ensuring that the result can be directly parsed by json.loads.
\#[Information]: 
\{information\}

\#example:
[\{\{"user": "I wanted to talk to you about my cousin, Ethan Parker. He's 39 years old.", "assistant": "Certainly! Ethan is 39. Is there something specific you'd like to discuss about him?"\}\},
\{\{"user": "Well, he's actually pretty short for a guy, only 162 cm. He always jokes about it, though.", "assistant": "It sounds like Ethan has a good sense of humor about his height. That's always a great quality!"\}\},
\{\{"user": "Yes, he really does. It's one of the things that makes him so fun to be around.", "assistant": "I can imagine! Having a cousin with a lighthearted attitude must make family gatherings enjoyable."\}\},
\{\{"user": "Definitely. And since we’re so close in age, it feels like he’s more of a friend than just family.", "assistant": "That sounds wonderful. It must be special to have that kind of bond with your cousin."\}\}
\{\{"user": "It really is. We’ve shared a lot of memories growing up together.", "assistant": "Those shared memories must make your relationship even stronger. It sounds like Ethan has been a big part of your life."\}\}]

\subsection{Observation Prompt}

\textbf{Role Message Prompt.} [User Message]: \{message\}
Please rewrite the above user message into a colloquial declarative sentence. Ensure it is smooth and free of grammatical errors, without changing the original information.
Only output the rewritten user message, without including the original message. Do not output any other description.
Output example: Lucas Grant, who is my boss, has a Master's degree.
"""

\textbf{Event Message Prompt.} [User Message]: \{message\}
Please rewrite the above user message into a colloquial declarative sentence. Ensure it is smooth and free of grammatical errors, without changing the original information, and avoid using 'you'. Don't forget use I , me or my 
Only output the rewritten user message, without including the original message. Do not output any other description.
Output example: Climb Fest draws a crowd of around nine hundred people.

\section{Result Details}

\subsection{Reflective Result}
In Tab~\ref{tab:10}, we show the detailed results of our 10k-Reflective memory dataset.
\begin{table*}[htbp]
  \caption{The results of different mechanisms on different types of our 10k-Reflective memory dataset.}
  \vspace{-0.2cm}
  \label{tab:10}%
\resizebox{\textwidth}{!}{
  \begin{tabular}{ccccc}
\hline
\hline
\multirow{2}[4]{*}{Method} & \multicolumn{2}{c}{Participation-Accuracy} & \multicolumn{2}{c}{Observation-Accuracy}
\bigstrut\\
\cline{2-5} & preference & emotion & preference & emotion \bigstrut\\
\hline
FullMemory & 0.733  & 0.593  & 0.883 & 0.630   \bigstrut[t]\\
RecentMemory & 0.700  & 0.481  & 0.867 & 0.556   \\
RetrievalMemmory & 0.692  & 0.556  & 0.883  & 0.593 \\
GenerativeAgent & 0.742  & 0.412  & 0.883  & 0.676     \\
MemoryBank & 0.692  & 0.296  & 0.900  & 0.481  \\
MemGPT & 0.733  & 0.471  & 0.883  & 0.556   \\
SCMemory & 0.542  & 0.294  & 0.783  & 0.333  \bigstrut[b]\\
\hline
\hline
\end{tabular}%
}
\vspace{-0.3cm}
\end{table*}%

\subsection{Facutal Result}

In Tab~\ref{tab:11}, we show the detailed results of our 10k-Factual-Partipation memory dataset.
In Tab~\ref{tab:12}, we show the detailed results of our 10k-Factual-Observation memory dataset.

\begin{table*}[htbp]
  \caption{The results of different mechanisms on different types of our 10k-Factual-Participation dataset. Including sh(sigle-hop), mh(multi-hop), comp(comparative), agg(aggregative), pp(post-processing), ku(knowledge-update), ssa(signle-session-assistant), msa(multi-session-assistant).}
  \vspace{-0.2cm}
  \label{tab:11}%
\resizebox{\textwidth}{!}{
  \begin{tabular}{ccccccccc}
\hline
\hline
\multirow{1}[4]{*}{Method} & \multicolumn{8}{c}{Participation-Accuracy}
\bigstrut\\
\cline{2-9}  & sh   & mh  & comp    & agg    & pp    & ku  & ssa    & msa \bigstrut\\
\hline
FullMemory & 0.825  & 0.8  & 0.55 & 0.275 & 0.625  & 0.75  & 0.7 & 0.55 \bigstrut[t]\\
RecentMemory & 0.85  & 0.75  & 0.425 & 0.45 & 0.65  & 0.725  & 0.717 & 0.5 \\
RetrievalMemmory & 0.875  & 0.775  & 0.55  & 0.275  & 0.475  & 0.675  & 0.4  & 0.3  \\
GenerativeAgent & 0.75  & 0.675  & 0.3  & 0.35  & 0.525  & 0.525  & 0.267  & 0.55  \\
MemoryBank & 0.575  & 0.7  & 0.25  & 0.25  & 0.475  & 0.55  & 0.417  & 0.4  \\
MemGPT & 0.625  & 0.625  & 0.275  & 0.225 & 0.45  & 0.625  & 0.367  & 0.45 \\
SCMemory & 0.575  & 0.475  & 0.05  & 0.275  & 0.525  & 0.475  & 0.217  & 0.1  \bigstrut[b]\\
\hline
\hline
\end{tabular}%
}
\vspace{-0.3cm}
\end{table*}%

\begin{table*}[htbp]
  \caption{The results of different mechanisms on different types of our 10k-Factual-Observation memory dataset. Including sh(sigle-hop), mh(multi-hop), comp(comparative), agg(aggregative), pp(post-processing), ku(knowledge-update).}
  \vspace{-0.2cm}
  \label{tab:12}%
\resizebox{\textwidth}{!}{
  \begin{tabular}{ccccccc}
\hline
\hline
\multirow{1}[4]{*}{Method} & \multicolumn{6}{c}{Observation-Accuracy}
\bigstrut\\
\cline{2-7}  & sh   & mh  & comp    & agg    & pp    & ku \bigstrut\\
\hline
FullMemory & 0.92  & 0.92  & 0.667 & 0.233 & 0.82  & 0.6   \bigstrut[t]\\
RecentMemory & 0.92  & 0.92  & 0.667 & 0.367 & 0.82  & 0.65   \\
RetrievalMemmory & 0.92  & 0.92  & 0.633  & 0.367  & 0.78  & 0.45    \\
GenerativeAgent & 0.88  & 0.94  & 0.7  & 0.3  & 0.82  & 0.4    \\
MemoryBank & 0.8  & 0.78  & 0.633  & 0.233  & 0.800  & 0.6    \\
MemGPT & 0.94  & 0.92  & 0.667  & 0.233 & 0.82  & 0.600   \\
SCMemory & 0.46  & 0.68  & 0.133  & 0.133  & 0.78  & 0.65    \bigstrut[b]\\
\hline
\hline
\end{tabular}%
}
\vspace{-0.3cm}
\end{table*}%

\end{document}